\documentclass[10pt,twocolumn,letterpaper]{article}

\usepackage{cvpr}              
\usepackage[accsupp]{axessibility}

\usepackage{graphicx}
\usepackage{amsmath}
\usepackage{amssymb}
\usepackage{booktabs}
\usepackage{multirow}
\usepackage{multicol}

\usepackage[utf8]{inputenc} 
\usepackage[T1]{fontenc}    
\usepackage{url}            
\usepackage{amsfonts}       
\usepackage{nicefrac}       
\usepackage{microtype}      
\usepackage{xcolor}         
\usepackage{subcaption}
\usepackage{xspace}
\usepackage{enumitem}
\usepackage{algorithm}
\usepackage{algpseudocode}
\usepackage{listings}
\usepackage{bm}
\usepackage{amsbsy}
\usepackage{bbding}

\newcommand\algcomment[1]{\def\@algcomment{\footnotesize#1}}

\newcommand{\tablestyle}[2]{\setlength{\tabcolsep}{#1}\renewcommand{\arraystretch}{#2}\centering\footnotesize}
\newcommand{\authorskip}{\hspace{3mm}}

\definecolor{citecolor}{HTML}{0071BC}
\definecolor{linkcolor}{HTML}{ED1C24}
\usepackage[pagebackref=true,breaklinks=true,letterpaper=true,colorlinks,bookmarks=false,citecolor=citecolor, linkcolor=linkcolor]{hyperref}

\usepackage[capitalize]{cleveref}
\crefname{section}{Sec.}{Secs.}
\Crefname{section}{Section}{Sections}
\Crefname{table}{Table}{Tables}
\crefname{table}{Tab.}{Tabs.}

\def\cvprPaperID{4248} 
\def\confName{CVPR}
\def\confYear{2023}

\begin{document}

\title{AltFreezing for More General Video Face Forgery Detection}

\author{Zhendong Wang{ $^{1,}$}\footnotemark[1] \authorskip Jianmin Bao{$^{2,}$}\footnotemark[1] \authorskip Wengang Zhou{$^{1,3,}$}\footnotemark[2] \authorskip
Weilun Wang{$^{1}$} \authorskip Houqiang Li{$^{1,3,}$}\footnotemark[2] \\
 $^{1}$ CAS Key Laboratory of GIPAS, EEIS Department, University of Science and Technology of China \\
$^{2}$ Microsoft Research Asia \\
$^{3}$ Institute of Artificial Intelligence, Hefei Comprehensive National Science Center \\
{\tt\small \{zhendongwang,wwlustc\}@mail.ustc.edu.cn } \\
{\tt\small jianbao@microsoft.com, \{zhwg,lihq\}@ustc.edu.cn}
}
\maketitle
\renewcommand{\thefootnote}{\fnsymbol{footnote}}
\footnotetext[1]{Equal contribution.} 
\footnotetext[2]{Corresponding authors.} 
\begin{abstract}

Existing face forgery detection models try to discriminate fake images by detecting only spatial artifacts~(\eg, generative artifacts, blending) or mainly temporal artifacts~(\eg, flickering, discontinuity). They may experience significant performance degradation when facing out-domain artifacts. In this paper, we propose to capture both spatial and temporal artifacts in one model for face forgery detection. A simple idea is to leverage a spatiotemporal model~(3D ConvNet). However, we find that it may easily rely on one type of artifact and ignore the other.
To address this issue, we present a novel training strategy called AltFreezing for more general face forgery detection. The AltFreezing aims to encourage the model to detect both spatial and temporal artifacts. It divides the weights of a spatiotemporal network into two groups: spatial-related and temporal-related. Then the two groups of weights are alternately frozen during the training process so that the model can learn spatial and temporal features to distinguish real or fake videos. Furthermore, we introduce various video-level data augmentation methods to improve the generalization capability of the forgery detection model. Extensive experiments show that our framework outperforms existing methods in terms of generalization  to unseen manipulations and datasets. Code is available at \url{https://github.com/ZhendongWang6/AltFreezing}.

\end{abstract}

\section{Introduction}

With the recent rapid development of face generation and manipulation techniques~\cite{vougioukas2019realistic, li2020advancing, thies2020neural, thies2019deferred, thies2016face2face, jiang2020deeperforensics, vougioukas2019end, zhou2019talking}, it has become very easy to modify and manipulate the identities or attributes given a face video. 
This brings many important and impressive applications for movie-making, funny video generation, and so on. 
However, these techniques can also be abused for malicious purposes, creating serious crisis of trust and security in our society. 
Therefore, how to detect video face forgeries has become a hot research topic recently.

\begin{figure}[t] 
    \centering 
    \includegraphics[width=0.75\linewidth]{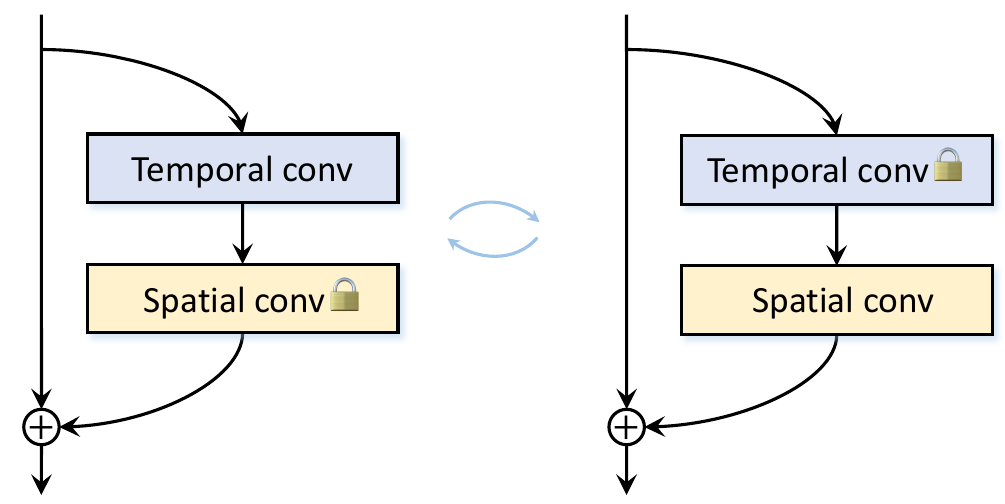}
    \caption{\textbf{Illustration of AltFreezing training strategy} in a building block of the spatiotemporal network. The convolutional kernels of the spatiotemporal network are divided into two groups: \emph{temporal-based} and \emph{spatial-based}. Two groups of weights are alternately frozen during training. With the help of the alternate freezing~(\emph{AltFreezing}) strategy, our model can capture both spatial and temporal artifacts to distinguish between fake and real videos.}
    \label{fig:AltFreezing}
    \vspace{-2.0mm}
\end{figure}

To date, one successful line of research~\cite{li2020face,sbi,roessler2019faceforensicspp,wang2020cnn,chai2020makes,nguyen2019multi,li2018exposing} tries to discriminate fake images by detecting ``spatial'' artifacts in the generated images (\eg, checkboard, unnaturalness, and characteristic artifacts underlying the generative model). While these methods achieve impressive results in searching spatial-related artifacts, they ignore  the temporal coherence of a video and fail to capture ``temporal'' artifacts like flicking and discontinuity in the video face forgeries. 
Some recent works~\cite{haliassos2021lips,FTCN,sabir2019recurrent} notice this issue and try to address it by leveraging temporal clues. 
Although they achieve encouraging results in detecting unnatural artifacts at the temporal level, the resulting models are not sufficiently capable of finding spatial-related artifacts.

In this paper, we attempt to capture both spatial and temporal 
artifacts for general video face forgery detection. 
Generally, a well-trained spatiotemporal network~(3D ConvNet) has the capability of searching both spatial and temporal artifacts.
However, we find that na\"ive training may cause it to easily rely on spatial artifacts while ignoring temporal artifacts to make a decision, causing a poor generalization capability. 
This is because spatial artifacts are usually more obvious than temporal incoherence, na\"ively optimizing a 3D convolutional network makes it easily rely on spatial artifacts.  

So the question is how to enable the spatiotemporal network to capture both spatial and temporal artifacts. To this end,  we propose a novel training strategy called \emph{AltFreezing}. As shown in \cref{fig:AltFreezing}, the key idea is to alternately freeze spatial- and temporal-related weights during training. Specifically, a spatiotemporal network~\cite{3dr50} is built upon 3D res-blocks, which consist of spatial convolution with kernel size as $1 \times K_h \times K_w$ and temporal convolution with kernel size as $K_t \times 1 \times 1$. 
These spatial and temporal convolutional kernels are responsible for capturing spatial- and temporal-level features, respectively. 
Our AltFreezing strategy encourages the two groups of weights to be updated alternately so that both spatial and temporal artifacts can be conquered. 

Furthermore, we propose a set of video-level fake video argumentation methods for generating fake videos for training. 
These methods could be divided into two groups. 
The first is fake clips that only involve temporal artifacts wherein we just randomly drop and repeat frames for real clips. 
The second is clips with only  spatial artifacts that are obtained by blending a region from one real clip to another real clip. 
These augmentation methods are the first to take the temporal dimension into consideration and generate spatial-only and temporal-only fake videos. 
With these augmentations, the spatiotemporal model is further encouraged to capture  both spatial and temporal artifacts.

Equipped with the above-mentioned two techniques, we achieve state-of-the-art performance in various challenging face forgery detection scenarios, including generalization capability to unseen forgeries, and robustness to various perturbations. 
We also provide a comprehensive analysis of our method to verify the effectiveness of our proposed framework.

Our main contributions are three-fold as follows.
\begin{itemize}
    \item We propose to explore both spatial and temporal artifacts for video face forgery detection. To achieve this, a novel training strategy called AltFreezing is proposed. 
    \item We introduce video-level fake data augmentation methods to encourage the model to capture a more general representation of different types of forgeries.
    \item Extensive experiments on five benchmark datasets including both cross-manipulation and cross-dataset evaluations demonstrate that the proposed method sets new state-of-the-art performance.
\end{itemize}

\section{Related Work}
In the past few years, face forgery detection has been an emerging research area with the fast development of generative models and manipulation techniques. In this section, we briefly revisit the development of face forgery detection.

\subsection{Image Face Forgery Detection}

Earlier face forgery detection methods~\cite{afchar2018mesonet,fridrich2012rich, bayar2016deep, hsu2018learning,rahmouni2017distinguishing,wang2020cnn,chai2020makes} mainly focus on spatial artifacts of manipulated images, and directly train a binary classifier based on CNN or MLP as the detector. Later Rossler \etal~\cite{roessler2019faceforensicspp} suggest that an unconstrained Xception~\cite{chollet2017xception} network can achieve an impressive performance. some works pay more attention to special types of artifacts, such as frequency~\cite{qian2020thinking,jeong2022frepgan,jeong2022bihpf,masi2020two}, blending artifacts~\cite{li2020face,sbi,chen2022self}, resolution difference~\cite{li2019exposing}, and so on. Moreover, there are some works~\cite{dang2020detection, Islam_2020_CVPR, Wu_2019_CVPR,bappy2019hybrid,haiwei2022exploring,nguyen2019multi} aiming to localize the forged regions and make a decision based on the predicted regions. A more recent work ICT~\cite{dong2020identity, ict} tries to leverage identity information for detecting face forgeries.

\subsection{Video Face Forgery Detection}

Recent works~\cite{mittal2020emotions,haliassos2021lips,sabir2019recurrent,li2018ictu,de2020deepfake,amerini2019deepfake,cozzolino2021id} start to take temporal cues into consideration for face forgery detection. 
CNN-GRU~\cite{sabir2019recurrent} employs a GRU module after CNN to introduce the temporal information. In~\cite{de2020deepfake,amerini2019deepfake}, a 3D ConvNet is directly trained to detect spatial and temporal artifacts. 
Some studies introduce prior knowledge to benefit video face forgery detection, such as eye blinking~\cite{li2018ictu}, lip motion~\cite{haliassos2021lips}, and emotion~\cite{mittal2020emotions}. 
Amerini \etal~\cite{amerini2019deepfake} suggest that predicting optical flow between frames helps deepfake detection. 

There are A part of works~\cite{gu2021spatiotemporal,haliassos2022leveraging,FTCN} which tend to focus on representation learning. STIL~\cite{gu2021spatiotemporal} considers both the spatial and temporal inconsistency and designs a spatio-temporal inconsistency Learning framework for deepfake video detection.
RealForensics~\cite{haliassos2022leveraging} introduces audio information and leverages self-supervised learning for representation learning.
A recent work FTCN~\cite{FTCN} explores directly training a fully temporal 3D ConvNets with an attached  temporal Transformer.
However, detecting without spatial information may harm the generalization capability. 
In this work, we aim to bring both spatial and temporal features for more general face forgery detection.

\subsection{Generalization to Unseen Manipulations}

With the rapid development of face generation and manipulation techniques, many previous face forgery detection methods~\cite{wang2020cnn,roessler2019faceforensicspp,chai2020makes} cannot well address unseen manipulations and datasets. 
Recent studies have noticed this challenge and focus on improving the generalization capability of the model. FWA~\cite{li2019exposing} targets the artifacts in affine face warping as the distinctive feature to detect the forgery.
Face X-ray~\cite{li2020face} proposes that detecting blending boundaries in images can make a general detector, which sets up a new paradigm of synthesizing images for generalizable face forgery detection.
SBI~\cite{sbi} inherits the detecting boundaries thought proposed by Face X-ray~\cite{li2020face} and suggests that blending from single pristine images is more suitable. Another work SLADD~\cite{chen2022self} proposes to dynamically synthesize forged images by adversarial learning.

Besides image-level face forgery detection, there are also works~\cite{haliassos2021lips,haliassos2022leveraging,gu2021spatiotemporal,FTCN} paying attention to video-level face forgery detection.
LipForensics~\cite{haliassos2021lips} uses a network pre-trained on a LipReading dataset~\cite{chung2016lip} and then makes a prediction based on the mouth region, which relies on audio data.
RealForensics~\cite{haliassos2022leveraging} also introduces audio information and leverages self-supervised learning to learn a better representation of forgery discrimination.
FTCN~\cite{FTCN} takes full advantage of temporal incoherence to detect the forged videos, based on an assumption that detecting forgeries in the temporal dimension is more general. 
In this work, we make no assumption or hypothesis. 
Instead, we design a novel training strategy to  make full use of spatial and temporal information to make a prediction without extra data.  

\subsection{Data Synthesis for Face Forgery Detection}

Synthesizing  or augmenting data is a classic method to improve the diversity and amount of training datasets in deep learning. 
In face forgery detection, several works start from the data  synthesis viewpoint to seek a more general detector. FWA~\cite{li2019exposing} proposes to synthesize fake data by blurring facial regions based on the assumption that current deepfake algorithms can only generate images of limited resolutions. 
Face X-ray~\cite{li2020face}, I2G~\cite{zhao2021learning}, SLADD~\cite{chen2022self}, and SBI~\cite{sbi} propose to synthesize fake images by blending two images based on the thought of most manipulated images may produce blending boundary artifacts. 
Although those blending artifact detection methods achieve promising performance on generalization experiments, until recently, there is not a very effective video-level data synthesis method in face forgery detection. 
In this work, we aim to design video-level data augmentation methods which are more suitable for encouraging spatiotemporal networks to learn better spatial and temporal representation.

\section{Method}


\begin{algorithm}[t]
\caption{Pseudocode of AltFreezing in Pytorch.}
\label{alg:AltFreezing}
\definecolor{codeblue}{rgb}{0.25,0.5,0.5}
\lstset{
  backgroundcolor=\color{white},
  basicstyle=\fontsize{7.2pt}{7.2pt}\ttfamily\selectfont,
  columns=fullflexible,
  breaklines=true,
  captionpos=b,
  commentstyle=\fontsize{7.2pt}{7.2pt}\color{codeblue},
  keywordstyle=\fontsize{7.2pt}{7.2pt},
}
\begin{lstlisting}[language=python]
# F: a 3D spatiotemporal network
# V, y: video clips, labels
# I_s, I_t: iterations of freezing spatial, temporal kernels

def st_optimizer(network):
    # splitting params into 
    # spatial-related and temporal-related
    params_s, params_t = st_split(network)
    # alternate optimizer
    return SGD(params_s,...), SGD(params_t,...)

count = 0
optim_s, optim_t = st_optimizer(F)
for V, y in loader: # load a minibatch
    optim_t.zero_grad() # zero gradient
    optim_s.zero_grad() # zero gradient
    V = aug(V) # random augmentation
    pred = F(V) # network prediction
    loss = CrossEntropyLoss(pred, y) # compute loss
    loss.backward() # compute gradient
    if count %(I_s+I_t)<I_s: # spatial freezing 
        optim_t.step() # temporal optimization
    else: # temporal freezing
        optim_s.step() # spatial optimization
    count+=1
\end{lstlisting}
\end{algorithm}

\subsection{Motivation}
Artifacts in forged face images can be roughly divided into two types: spatial-related~(\eg, generative artifacts, blending, and \emph{etc.}) and temporal-related artifacts (\eg, flickering and discontinuity).
Earlier works~\cite{bayar2016deep,chollet2017xception,chai2020makes} mostly focus on searching spatial artifacts. These artifacts can be easily captured by training a deep neural network. However, these image-level face forgery detection methods do not have the capability of capturing temporal-level artifacts.


With the demand for detecting more challenging forgeries, research on how to detect video-level face forgeries attracts more and more attention. Researchers seek to leverage video-level artifacts for detecting fake videos. Among them, the typical works are LipForensics~\cite{haliassos2021lips} and FTCN~\cite{FTCN}. They achieve impressive results in detecting temporal-level artifacts like unnatural lip motion or temporal incoherence.
However, they have a strong assumption that focusing on temporal incoherence contributes to a more general detector. Indeed, currently most face manipulation and generation methods~\cite{li2020advancing, thies2020neural, thies2019deferred, thies2016face2face, vougioukas2019realistic, vougioukas2019end,zhu2021one} generate forged videos in a frame-by-frame manner, yielding flicking and incoherent artifacts. However, few but not none, there are also video-level manipulation and generation methods~\cite{alaluf2022third,fu2022m3l,xu2022mobilefaceswap}, which can produce videos that are more coherent perceptually, making these temporal-based detectors difficult to handle. 
On the other hand, these generative methods still contain spatial-level artifacts to some extent. Hence, it is urgent to develop a general face forgery detector for capturing both spatial and temporal artifacts.


A spatial-temporal network is theoretically capable of capturing both spatial and temporal artifacts. However, we observe that if we na\"ively train a spatiotemporal network with a binary classification loss, the network will rely on one type of ``easy'' artifact to distinguish real or fake. It makes the detector cannot completely to find all the artifacts for classification. This will cause the detector to have a terrible generalization capability to unseen deepfake datasets or manipulation methods.


To address this issue, we propose a novel training strategy named \textit{AltFreezing}, to encourage the model to capture both spatial and temporal artifacts. We assume that capturing both the spatial- and temporal-level artifacts can yield a strong generalization capability for unseen datasets and manipulation methods.

Moreover, we notice that data augmentation plays an increasingly important role in improving the generalization capability of face forgery detection. However, most of them are at the image level. To encourage the spatiotemporal network to encompass a strong generalization capability, we further introduce some video-level augmentation techniques.

\subsection{AltFreezing}
Our AltFreezing is a simple modification to the standard spatiotemporal network updating mechanism. It first divides the weights of the network into two groups. Then the two groups of weights are alternately frozen during the training process. In other words, AltFreezing updates the weights of two groups in turn.
Take a typical 3D ConvNet, 3D ResNet-50~(R50)~\cite{3dr50} as an example.
The convolutional weights of 3D R50 can be mainly divided into spatial-based~(\ie, $1\times K_h\times K_w$ convolutional kernels) and temporal-based~(\ie, $K_t\times 1\times 1$ convolutional kernels). Note that, $1\times 1\times 1$ convolutional layers, linear layers, batch normalization layers, and other modules with parameters are regarded as both related considering these layers do not have a receptive field on both temporal and spatial dimensions. After splitting, AltFreezing   starts to freeze the two groups of weights alternately during the training stage. Specifically, when the spatial-related weights are frozen, the network will strive to search temporal artifacts to distinguish between real and fake. Similarly, when the temporal-related weights are frozen, the network will struggle to search spatial artifacts to discriminate between real and fake.

\begin{figure}[t] 
    \centering 
    \includegraphics[width=1.0\linewidth]{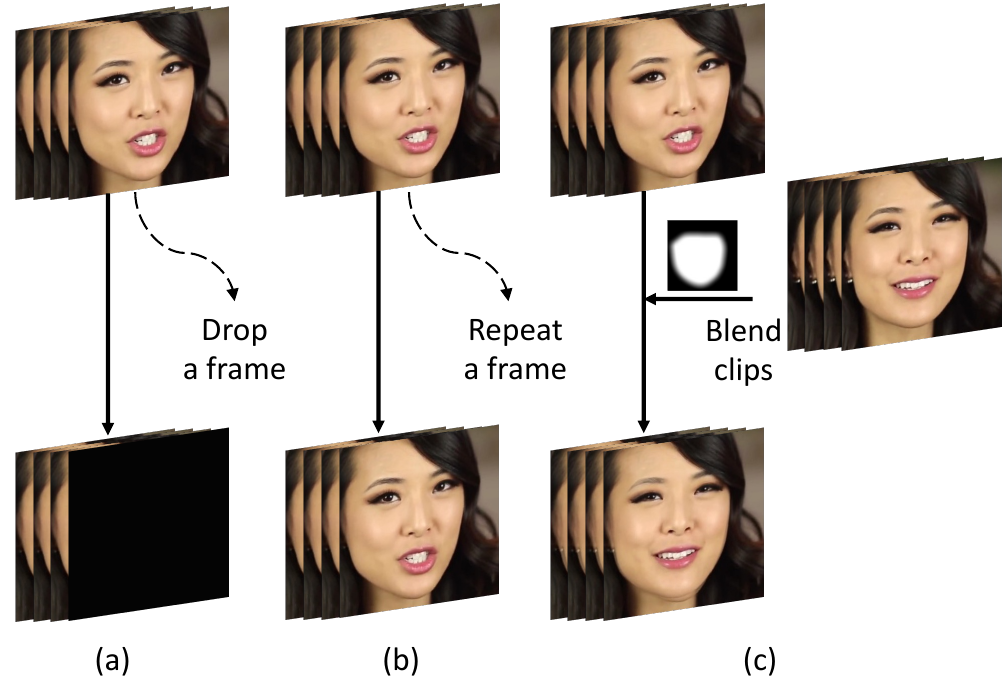}
    \vspace{-2em}
    \caption{\textbf{Illustration the proposed Fake Clip Generation.} For each video clip during training, we randomly use a) temporal dropout, b) temporal repeat, and c) clip-level blending to generate a fake clip for generating fake samples. Temporal dropout and repeat can introduce fake clips with challenging temporal incoherence. Clip-level blending can generate fake clips which only contain challenging spatial artifacts.}
    \label{fig:video_aug}
\end{figure}

Suppose that the weights $\theta$ of the spatiotemporal network $F$ are divided into $\theta_S$ and $\theta_T$. Given training data~(input video clips $V$, real/fake labels $y$), our goal of training a video face forgery detector is to minimize the loss function $\mathcal{L}(\mathbf{F}(V;\theta_S, \theta_T), y)$ by optimizing the weights $\theta_S$ and $\theta_T$ of spatiotemporal network $\mathbf{F}$. 
The update of $\theta_S$  is formulated as follows, 
\begin{equation}
    \theta_S \leftarrow \theta_S - \alpha \frac{\partial \mathcal{L}(\mathbf{F}(I;\theta_S, \theta_T), y)}{\partial \theta_S},
\end{equation}
where $\alpha$ is the learning rate. 
Correspondingly, the update of $\theta_T$ is formulated as follows,
\begin{equation}
    \theta_T \leftarrow \theta_T - \alpha \frac{\partial \mathcal{L}(\mathbf{F}(I;\theta_S, \theta_T), y)}{\partial \theta_T}.
\end{equation}

Moreover, we can control the ratio of iterations in freezing spatial weights and temporal weights $I_s$:$I_t$ to encourage the network to pay more attention to spatial or temporal artifacts. Within a cycle, we first freeze the spatial weights $I_s$ iterations then we freeze the temporal weights $I_t$ iterations. We find that spatial artifacts are usually easy to learn, so $I_s$ is set to be larger than $I_t$. The whole algorithm of AltFreezing is summarised in \cref{alg:AltFreezing}. With the help of the alternately freezing strategy, the network cannot easily converge by only focusing on one single type of artifact. By switching between spatial and temporal weights, the final trained network is enabled with an ability to capture both spatial and temporal artifacts for more general face forgery detection.

\subsection{Fake Clip Generation}
\label{subsec: video_aug}
Some recent methods~\cite{das2021towards,li2019exposing, li2020face,sbi} leverage data augmentations to encourage a more general representation learning for detecting face forgeries. However, these augmentations are only at the image level. Until recently in the face forgery detection area, little attention is paid to video-level augmentations which are actually more compatible with 3D ConvNets. To learn better video-level representation,  we propose a set of fake video synthetic methods including temporal-level and spatial-level augmentations. 

As shown in \cref{fig:video_aug}, we propose three video-based augmentations, \ie, temporal dropout, temporal repeat, and clip-level blending. The first two types are temporal-related fake clip generation. Temporal dropout~(\cref{fig:video_aug}~(a)) means one or multiple random frames of the video clip are dropped, which is a strong imitation of cutting video frames. After dropping, frames after the dropped frames are shifted forward, and the empty frames are set to 0. For performing temporal repeat~(\cref{fig:video_aug}~(b)), one or multiple random frames are repeated, which is a strong simulation of inserting frames into an original video. Then the frames after the repeated frames are shifted backward, and the extra frames are removed. These two temporal-based augmentations can help the spatiotemporal network to capture temporal artifacts.


On the contrary, the proposed clip-level blending~(\cref{fig:video_aug}~(c)) is spatial-related fake clip generation.
Concretely, we first randomly choose two clips from a single video or two videos, in which one serves as the foreground clip $V_f$ and the other serves as the background clip $V_b$, 
After that, we generate a random mask $M$ delimiting the manipulated region, with each pixel having a greyscale value between 0.0 and 1.0. Then we use the mask to blend each frame from the foreground clip to its corresponding frame of the background clip by: 
\begin{equation}
\label{eqn:blending}
V^i = V^i_f*M+V^i_b*(1-M), 
\end{equation}
where $i = 1,2,\cdots,L$ is the $i$-th frame of the clip, $L$ is the length of the clip. Since $V_f$ and $V_b$ are real clips that are temporally coherent, the resulting clip $V$ is also temporally coherent since the blend operation does not corrupt temporal coherence. Thus $V$ only contains spatial-related artifacts, \ie, the blending boundary around the  mask $M$. Our method is different from the previous image-level blending methods~\cite{li2020face, sbi,chen2022self}, which process each image independently yielding temporal incoherence.

Incorporating the video-level fake clip augmentations with AltFreezing,  our spatiotemporal network can capture more general spatial and temporal artifacts for face forgery detection.
Finally, our model is trained  with a simple binary cross-entropy loss, which is formulated as follows, 
\begin{equation}
    \mathcal{L}(\widetilde{y}, y)= -\sum_{i=1}^N (y^i*\mathrm{log}(\widetilde{y}^i)+(1-y^i)*\mathrm{log}(1-\widetilde{y}^i)),
\end{equation}
where $N$ denotes mini-batch size, $y$ is the label, and $\widetilde{y}$ is the prediction of the network.

\section{Experiment}
\subsection{Setup}

\noindent\textbf{Datasets.} (1) \textbf{FaceForensics++}~(FF++)~\cite{roessler2019faceforensicspp} consists of 1,000 real videos and 4,000 fake videos. The fake videos are generated by four manipulation methods~(Deepfake~(DF)~\cite{deepfakes}, Face2Face~(F2F)~\cite{thies2016face2face}, FaceSwap~(FS)~\cite{faceswap},  NeuralTexture(NT)~\cite{thies2020neural}). (2) \textbf{CelebDF v2}~(CDF)~\cite{Celeb_DF_cvpr20} is a new face-swapping dataset including 5,639 synthetic videos and 890 real videos downloaded from YouTube. In our experiments, its 518 testing videos are used for evaluation. (3) \textbf{Deepfake Detection}~(DFD)~\cite{deepfakedetection}
contains over 3,000 manipulated videos from 28 actors in various scenes. (4) \textbf{FaceShifter}~(FSh)~\cite{li2020advancing} and \textbf{DeeperForensics}~(DFo)~\cite{jiang2020deeperforensics} generate high-fidelity face-swapping videos based on the real videos from FF++. In our experiments, we use the training split of FF++ as the training data by default. Unless stated otherwise, we use the c23 version of FF++, following recent literatures~\cite{FTCN,haliassos2021lips}.

\noindent\textbf{Evaluation Metrics.} Following previous methods~\cite{li2020face,FTCN,sbi,haliassos2021lips}, we report the area under the receiver operating characteristic curve~(AUC) to evaluate the performance. Following \cite{FTCN,haliassos2021lips}, we report video-level AUC for fair comparisons. And for image-based methods, we average the frame-level predictions as the corresponding video-level prediction.

\noindent\textbf{Implementation details.} We use 3D ResNet50~\cite{3dr50} as our network and train it for 1k epochs with the SGD optimizer. The batch size is 32. We random sample concussive 32 frames of each video during training. The initial learning rate is 0.05, decayed to 0 at the ending epoch following the curve of cosine annealing. For data augmentations, RandomHorizontalFlip, RandomCutOut, and AddGaussianNoise are also applied besides the proposed video-level fake video synthetic augmentations.

\begin{figure*}[t]
	\includegraphics[width=\linewidth]{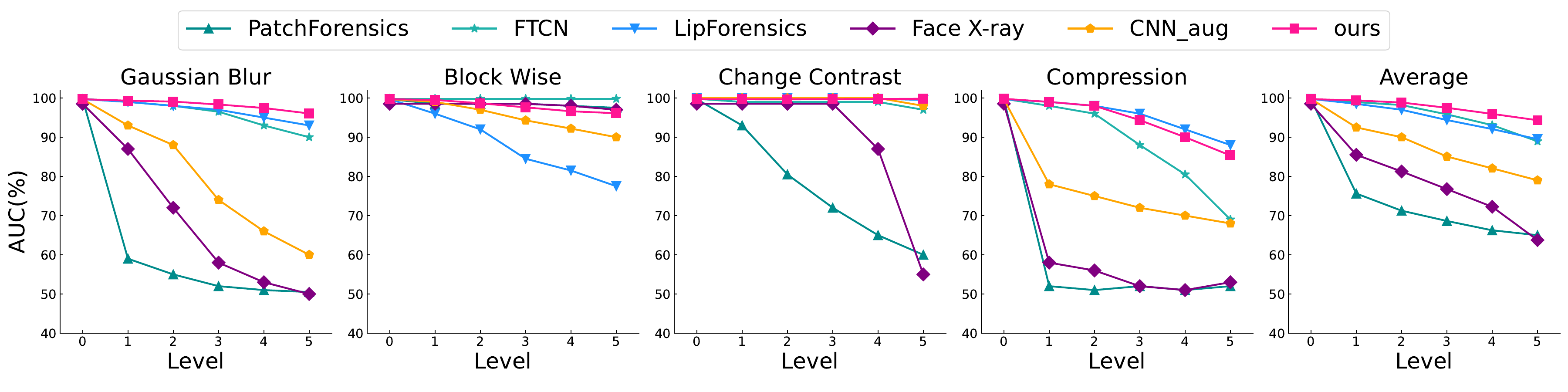}
\vspace{-2.5em}
\caption{\textbf{Robustness to unseen perturbations.} Video-level AUC(\%) is reported under five different degradation levels of four particular types of perturbations~\cite{jiang2020deeperforensics}. ``Average'' AUC score denotes the mean of each perturbation at each level.}
\label{fig:rubostness}
\vspace{-1mm}
\end{figure*}

\begin{table}[t]
    \centering
    \resizebox{1.0\linewidth}{!}{
    \begin{tabular}{lccccc}
\hline
Method  & CDF  & DFD   & FSh   & DFo   & Avg \\ \hline
Xception~\cite{roessler2019faceforensicspp} & 73.7 & --   & 72.0  & 84.5  & -- \\
CNN-aug~\cite{wang2020cnn} & 75.6 & -- & 65.7 & 74.4  & -- \\
PatchForensics~\cite{chai2020makes}  & 69.6 & --  & 57.8 & 81.8  & --  \\
Multi-task~\cite{nguyen2019multi}  & 75.7 & --  & 66.0  & 77.7  & --  \\
 FWA~\cite{li2018exposing}  & 69.5 & -- & 65.5  & 50.2 & -- \\
Two-branch~\cite{masi2020two}   & 76.7 & --   & --  & -- & -- \\
Face X-ray~\cite{li2020face}   & 79.5 & 95.4  & 92.8 & 86.8 & 88.6  \\
SLADD~\cite{chen2022self} & 79.7 & -- & -- & -- & -- \\
SBI-R50*~\cite{sbi} &  85.7 & 94.0 & 78.2 & 91.4 & 87.3 \\\hline
CNN-GRU~\cite{sabir2019recurrent}  & 69.8 & --  & 80.8 & 74.1  & --  \\
STIL~\cite{gu2021spatiotemporal} & 75.6 & -- & -- & -- & -- \\
LipForensics-Scratch~\cite{haliassos2021lips} & 62.5 & --  & 84.7  & 84.8 & --  \\ 
LipForensics~\cite{haliassos2021lips} & 82.4 & --  & 97.1  & 97.6 & --  \\ 
RealForensics-Scratch~\cite{haliassos2022leveraging} & 69.4 & -- & 87.9 & 89.3 & -- \\
RealForensics~\cite{haliassos2022leveraging} & 86.9 & -- & \textbf{99.7} & \textbf{99.3} & --  \\
FTCN~\cite{FTCN}  & 86.9 & 94.4 & 98.8  & 98.8  & 94.7 \\  
 \hline
\textbf{AltFreezing~(ours)} & \textbf{89.5} & \textbf{98.5} & 99.4 & \textbf{99.3} & \textbf{96.7} \\ \hline
\end{tabular}}
\vspace{-2mm}
\caption{\textbf{Generalization to unseen datasets.} We report the video-level AUC~(\%) on four unseen datasets: Celeb-DF-v2~(CDF), DeepFake Detection~(DFD), FaceShifter~(FSh), and DeeperForensics~(DFo). The models are trained on FF++ and tested on these unseen datasets. * denotes our reproduction with the official code, due to its unfair experiments using the raw version of training data of FF++. The results of other methods are from \cite{haliassos2021lips}.}
\label{tab:cross_dataset}
\end{table}

\begin{table}[t]
\centering
\resizebox{1.0\columnwidth}{!}{
\setlength\tabcolsep{2pt}
\begin{tabular}{lccccc}
\hline
Model & \#params & Arch & FSh & DFo &  Avg \\ \hline
LipForensics-Scratch~\cite{haliassos2021lips}  & 36.0M & R18+MS-TCN & 84.7  & 84.8 & 84.8 \\ 
LipForensics~\cite{haliassos2021lips}  & 36.0M & R18+MS-TCN & 97.1  & 97.6 & 97.4 \\ 
FTCN~\cite{FTCN} & \textbf{26.6M} & 3D R50+TT & 98.8  & 98.8  & 98.8 \\ \hline
\textbf{AltFreezing~(ours)}  & 27.2M & 3D R50  & \textbf{99.4} & \textbf{99.3} & \textbf{99.4}  \\ \hline
\end{tabular}
}
\vspace{-2mm}
\caption{\textbf{Comparison with video-level state-of-the-art methods in terms of parameters and architectures.} Video-level AUC(\%) is reported on FSh and DFo using models trained on FF++~\cite{roessler2019faceforensicspp}. ``MS-TCN" means multi-scale temporal convolutional network. ``TT" means temporal Transformer. Note that, 3D R50 used in FTCN~\cite{FTCN} is without spatial kernels.}
\label{table:compare_params}
\end{table}

\subsection{Generalization to Unseen Datasets}
In real-world scenarios, there is usually a gap between the tested forged videos and fake videos from the training dataset. Therefore, the generalization capability of models to unseen datasets is critical. To evaluate the generalization ability of our model, we use the original videos and all four types of fake videos in FF++~\cite{roessler2019faceforensicspp} as the training data, then evaluate the performance on CDF~\cite{Celeb_DF_cvpr20}, DFD~\cite{deepfakedetection}, FSh~\cite{li2020advancing}, and DFo~\cite{jiang2020deeperforensics}. 

We report the AUC results in \cref{tab:cross_dataset}. We observe that our model achieves the best performance on  CDF~(89.5\%), DFD~(98.5\%), and DFo~(99.3\%), and competitive performance on FSh~(99.4\%). It is worth noting that all other methods perform unsatisfactorily on CDF, while our model obtains an AUC of 89.5\%, which shows our model's strong generalization capability. And observing from the average AUC comparison, our method achieves a significant improvement of 2\% AUC score compared to previous video-level state-of-the-art method~\cite{FTCN}, 94.7\% $\rightarrow$ 96.7\%.
Moreover, we also compare the parameters and architectures in \cref{table:compare_params}. We observe that our method achieves the best performance with a simple 3D R50 network, which further indicates that our model is a simple but more general face forgery detector.

\subsection{Generalization to Unseen Manipulations}
For a general face forgery detector, they usually do not know which manipulation is applied to the tested videos. It is important to have a strong generalization capability to unseen manipulations. Following previous works~\cite{FTCN,haliassos2021lips}, we conduct the experiments on FF++~\cite{roessler2019faceforensicspp} with a leave-one-out setting. There are four types of forged face videos, \ie, DF, F2F, FS, and NT in FF++. We choose three of the forgery subsets as the training set. The remaining subset is used for evaluating the generalization capability of the model. 

In \cref{tab:cross-manipulation}, we show comparisons of our method with other state-of-the-art methods. The AUC scores demonstrate that our model can achieve impressive performance on the whole FF++ test set~(average AUC: 98.6\%), especially on the subsets DF~(99.8\%) and FS~(99.7\%) compared to previous methods. On F2F and NT, our results of ours are slightly lower than LipForensics~\cite{haliassos2021lips}. One possible explanation is that LipForensics~\cite{haliassos2021lips} employs a  pre-trained model with a strong prior knowledge of the mouth region, which is beneficial for unseen manipulation detection. While our model is trained from scratch, without using pre-trained models. Nonetheless, our model achieves better performance over LipForensics in terms of the average AUC on the four types of manipulation methods.

\begin{table}[t]
\footnotesize
\centering
\resizebox{1.0\columnwidth}{!}{
\begin{tabular}{lccccc}

\hline
\multirow{2}{*}{Method} & \multicolumn{4}{c}{Train on remaining three} &      \\ \cline{2-5}
                        & DF        & FS        & F2F       & NT       & Avg  \\ \hline
Xception~\cite{roessler2019faceforensicspp}              & 93.9      & 51.2      & 86.8      & 79.7     & 77.9 \\
CNN-aug~\cite{wang2020cnn}              & 87.5      & 56.3      & 80.1      & 67.8     & 72.9 \\
PatchForensics~\cite{chai2020makes}             & 94.0      & 60.5      & 87.3      & 84.8     & 81.7 \\
Face X-ray~\cite{li2020face}             & 99.5      & 93.2      & 94.5      & 92.5     & 94.9 \\ \hline
CNN-GRU~\cite{sabir2019recurrent}              & 97.6      & 47.6      & 85.8      & 86.6     & 79.4 \\
LipForensics-Scratch~\cite{haliassos2021lips}& 93.0      & 56.7      & 98.8      & 98.3     & 86.7 \\
LipForensics~\cite{haliassos2021lips}            & 99.7      & 90.1      & \textbf{99.7}      & \textbf{99.1}     & 97.1 \\
FTCN*~\cite{FTCN}  &   \textbf{99.8}       &  99.6         & 98.2         &  95.6  &   98.3  \\ \hline
\textbf{AltFreezing~(ours)} & \textbf{99.8} & \textbf{99.7} & 98.6 & 96.2 & \textbf{98.6} \\\hline

\end{tabular}}
\vspace{-2mm}
\caption{\textbf{Generalization to unseen manipulations.} We report the video-level AUC~(\%) on the FF++ dataset, which consists of four manipulation methods~(DF, FS, F2F, NT). The experiments obey the leave-one-out rule as \cite{haliassos2021lips,FTCN}. The three subsets of fake videos are used as the training data, the other one serves as the testing data. * denotes our reproduction without a temporal Transformer. The results of other methods are from \cite{haliassos2021lips}.}
\label{tab:cross-manipulation} 
\end{table}

\subsection{Robustness to Unseen Perturbations}
Besides the generalization to unseen datasets and manipulations, the robustness to unseen perturbations is also a concerning problem in real-world scenes. Following \cite{jiang2020deeperforensics}, we evaluate the robustness of our model to unseen perturbations considering four different degradation types, \ie, Gaussian blur, Block-wise distortion, Change contrast, and Video compression. Each perturbation is operated at five levels to evaluate the robustness of models under different-level different-type distortion. We show the AUC results on these unseen perturbations in \cref{fig:rubostness}, using the model trained on FF++. We observe that our method outperforms previous methods a lot at every level on average, which indicates that our method is much more robust and generalizable. Especially on serious degradations~(level 4, 5 in \cref{fig:rubostness}), the AUC of our model surpasses others a lot, \ie, about 3\% improvement on level 4 and 4\% improvement on level 5. 


\subsection{Ablation Studies}
In this section, we perform ablation studies to verify the effectiveness of the proposed AltFreezing. We do not utilize fake clip generation techniques without special notations. All the models are trained on FF++~\cite{roessler2019faceforensicspp}, and tested in FF++~\cite{roessler2019faceforensicspp}, CelebDF v2~\cite{Celeb_DF_cvpr20}, and FaceShifter~\cite{li2020advancing} to evaluate the generalization capability of the models. 

\begin{table}[t]
\footnotesize
\centering
\resizebox{1.0\columnwidth}{!}{
\begin{tabular}{lccccc}
\hline
\multirow{2}{*}{Model} & \multicolumn{3}{c}{Train on FF++} &\\ \cline{2-4} 
& FF++ & CDF & FSh & Avg\\ \hline  
 3D R50 & 99.3   & 81.8   & 99.2  &  93.4   \\
3D R50~(freeze S. kernels) & 99.5 & 76.8 & 98.9 & 91.7\\
3D R50~(freeze T. kernels) & 99.4 & 80.6 & \textbf{99.4}& 93.1 \\
 \textbf{3D R50~(AltFreezing)} & \textbf{99.7} & \textbf{86.4} & 99.3& \textbf{95.1}
 \\ \hline
\end{tabular}
}
\vspace{-2mm}
\caption{\textbf{Ablation study of variants of AltFreezing.} Video-level AUC(\%) is reported. ``S." means spatial and ``T." means temporal.}
\label{tab:altfreezing}
\end{table}

\noindent\textbf{Effect of AltFreezing.} We design several variants of AltFreezing, We use the 3D Resnet50~\cite{3dr50}~(3D R50 for short) as the basic network structure. 1) a vanilla 3D R50 network without any change of network structure or training strategy. 
2) In 3D R50~(freeze S. kernels), we split the weights of 3D R50 into spatial-related and temporal-related as AltFreezing. And during training, we fix all the spatial kernels and only update the weights of temporal kernels. 3) Similar to 2), in 3D R50~(freeze T. kernels), we only update the weights of spatial kernels. 4) the proposed AltFreezing with Resnet50 as the backbone.

We report the AUC results of these models in \cref{tab:altfreezing}. Compared with the 3D R50 baseline, our proposed AltFreezing training strategy can significantly improve the performance of in-domain face forgery detection and out-domain face forgery detection. We also notice that simply freezing the spatial or temporal weights of the 3D R50 network can not obtain a performance gain and even damage the performance. So the key design is alternately freezing the spatial and temporal weights during training. Moreover, AltFreezing is a plug-and-play training strategy for capturing both spatial and temporal artifacts, which does not introduce any extra computation or parameters.


\begin{table}[t]
\footnotesize
\centering
\vspace{-1em}
\resizebox{1.0\columnwidth}{!}{
\setlength\tabcolsep{4pt}
\begin{tabular}{lccc}
\hline
\multirow{2}{*}{Model} & \multicolumn{2}{c}{Train on FF++} &\\ \cline{2-3} 
& Temporal Set & Spatial Set & Avg\\ \hline  
FTCN~\cite{FTCN}  & 74.8 & 75.8 & 75.3 \\
 3D R50  & 76.5   & 71.5 & 74.0 \\
 \textbf{3D R50~(AltFreezing)} & \textbf{80.6} & \textbf{84.5} & \textbf{82.6}
 \\ \hline
\end{tabular}
}
\vspace{-2mm}
\caption{\textbf{Effect of AltFreezing when facing more hard cases.} Video-level AUC(\%) is reported on our synthetic datasets. The temporal Set is a synthetic dataset with only temporal incoherence. Spatial Set is a synthetic dataset with only spatial artifacts. The models are all trained on FF++~\cite{roessler2019faceforensicspp}.}
\label{tab:synthetic}
\end{table}


\noindent\textbf{Does AltFreezing really learn how to capture spatial and temporal artifacts?} Although our motivation for AltFreezing is learning to search both spatial and temporal artifacts, the readers might wonder whether our AltFreezing can really achieve this goal. We conduct experiments in more challenging scenes to verify the ability of our model to capture spatial and temporal artifacts. We build two new challenging datasets based on the testing set of real data in FF++\cite{roessler2019faceforensicspp}. 1) Temporal Set: we  aim to build a test set that only contains temporal-related artifacts, we randomly drop or repeat frames from a real video clip that all frames are real with only temporal incoherence introduced. 2) Spatial Set: we aim to build a test set that only contains spatial-related artifacts, we use a random mask to extract all the same region from a clip, then blend the region back into the other clip, since each pixel of these two clips is coherent, the newly generated clip is temporal coherent with only spatial artifacts. It is worth noting that we do not use the proposed fake video augmentation methods described in \cref{subsec: video_aug} in this experiment, in order to evaluate the performance of the proposed AltFreezing training strategy. 

The evaluation results on these two hand-crafted datasets are reported in \cref{tab:synthetic}. Even though our model does not see any types of artifacts in the Temporal Set and Spatial Set during the training stage. It achieves strong performance on these test sets. Compared with FTCN, which is specially designed for detecting temporal artifacts, our AltFreezing achieves a better performance in detecting temporal artifacts. This validates that spatial convolution is also important for detecting temporal artifacts, restricting all the spatial kernels to 1 is not an optimal choice.

\begin{table}[t]
\footnotesize
\centering
\resizebox{\columnwidth}{!}{
\setlength\tabcolsep{12pt}
\begin{tabular}{ccccc}
\hline
Freezing & \multicolumn{3}{c}{Train on FF++} & \\ \cline{2-4} 
 ratio ($I_s:I_t$) &  FF++ & CDF & FSh & Avg\\ \hline  
 baseline & 99.3   & 81.8   & 99.2  &  93.4   \\ \hline
1:1 & 99.6 & 82.4 & 99.2 & 93.7  \\ 
 5:1 & 99.5   & 82.8   & \textbf{99.7}   & 94.0    \\
 10:1  & 99.6 & 83.4 & 99.1 & 94.0 \\
\textbf{20:1} & \textbf{99.7} & \textbf{86.4} & 99.3 & \textbf{95.1} \\
 30:1 & 99.5 & 82.1 & 99.2 & 93.6 \\ \hline
\end{tabular}
}
\vspace{-2mm}
\caption{\textbf{Ablation study of the freezing ratio of AltFreezing.} Video-level AUC(\%) is reported. ``baseline'' means a 3D R50 with end-to-end training.}
\vspace{-1em}
\label{tab:alterratio}
\end{table}

\begin{table}[t]
\footnotesize
\centering
\resizebox{\columnwidth}{!}{
\setlength\tabcolsep{14pt}
\begin{tabular}{ccccc}
\hline
Aug. & \multicolumn{3}{c}{Train on FF++} & \\ \cline{2-4} 
level &  FF++ & CDF & FSh & Avg\\ \hline   
 none & \textbf{99.7} & 86.4 & 99.3 & 95.1
 \\  
 image & 99.6 & 78.6 & \textbf{99.4} & 92.5
 \\       
 \textbf{video}& \textbf{99.7} &  \textbf{89.5} & 99.3 & \textbf{96.2}
 \\ \hline
\end{tabular}
}
\vspace{-2mm}
\caption{\textbf{image-level augmentation~\cite{sbi} \vs video-level augmentation.} Video-level AUC(\%) is reported. ``Aug.'' means augmentation. ``image'' level augmentation means that the blending used is image level like~\cite{li2020face,sbi} while keeping other augmentations unchanged. ``video'' level augmentation means that using the proposed video-level augmentation methods.}
\label{tab:video_aug}
\vspace{-1em}
\end{table}

\begin{table}[t]
\footnotesize
\centering
\resizebox{1.0\columnwidth}{!}{
\setlength\tabcolsep{14pt}
\begin{tabular}{lccccc}
\hline
\multirow{2}{*}{Backbone} & \multicolumn{3}{c}{Train on FF++} &\\ \cline{2-4} 
& FF++ & CDF & FSh & Avg\\ \hline  
 3D R50 & \textbf{99.7} & 89.5 & 99.3& 96.2 \\
 3D R101 & 99.6 & \textbf{90.4} & \textbf{99.4} & \textbf{96.5}
 \\ \hline
\end{tabular}
}
\vspace{-3mm}
\caption{\textbf{Different backbone architectures.} Video-level AUC~(\%) is reported. The models are trained on FF++.}
\vspace{-4mm}
\label{tab:backbone}
\end{table}

\begin{figure}[t] 
    \centering 
    \includegraphics[width=0.93\linewidth]{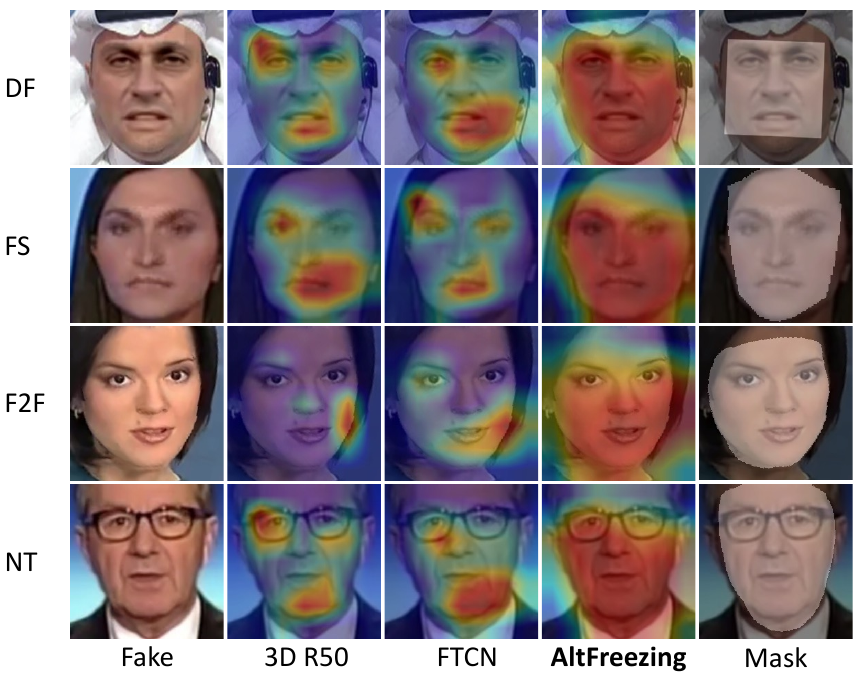}
    \vspace{-2mm}
    \caption{\textbf{Visualization of activation maps for fake samples from different manipulation methods.} Warmer color indicates a higher probability of fake. We compare vanilla 3D R50, FTCN, and 3D R50 with our AltFreezing strategy. The activation maps shown here are the mixing of activation heatmaps and the input fake frames. Our AltFreezing could locate the forgery region precisely.}
    \label{fig:cam}
\end{figure}
\noindent\textbf{Influence of the freezing ratio in AltFreezing.} In the AltFreezing algorithm, the ratio of iterations in freezing spatial weights and temporal weights $I_s$:$I_t$ controls the capability of the network on handling spatial-related and temporal-related artifacts. We conduct a comprehensive study about the effect of the freezing iterations ratio. We consider the freezing ratio in the set \{1:1, 5:1, 10:1, 20:1, 30:1\}, and train our models with a  different freezing ratio while keeping other configurations the same. 

\cref{tab:alterratio} shows that AltFreezing's performance initially increases and then decreases as the freezing ratio varies from 1:1 to 30:1. The model achieves the best generalization capability when the freezing ratio is 20:1. It is worth noting that AltFreezing is better than baseline (without AltFreezing) on the generalization ability of the model at all freezing ratios. Recent works FTCN~\cite{FTCN} and LipForiensics~\cite{haliassos2021lips} suggest that temporal information is more important for 3D networks to make a prediction. And combined with our analysis, spatial information may serve as a complement to temporal information for detecting face forgeries. So in our experiments, the freezing iterations of spatial-based kernels are more than temporal-based ones for the 3D ConvNet to capture more temporal artifacts and also involve spatial information.

\noindent\textbf{Effect of the fake clip generation.} The proposed fake clip generation method is based on three video-level augmentations to encourage a more general representation learning of video-level forgeries. Here, we study the effect of the data synthesis method.  As shown in \cref{tab:video_aug}, our video-level augmentations bring an average AUC improved from 95.1~\% to 96.2~\%. Especially on CDF~\cite{Celeb_DF_cvpr20}, with our fake video augmentations, our AltFreezing method gets a +3.1~\% 
absolute boost, 86.4\% $\rightarrow$ 89.5\%. This indicates that the proposed video-level fake sample synthesis benefits the generalization ability of the network a lot. We also compare our augmentations with recent self-blending image augmentation~\cite{sbi}. We find frame-level augmentation is not suitable to be directly applied to a spatiotemporal network for general face forgery detection. For more detailed experiments and discussions of fake clip 
generation, please refer to the supplemental material.

\noindent \textbf{Results of advanced architectures.} We further conduct an experiment about the effect of more advanced network architectures, as shown in \cref{tab:backbone}. Using 3D R101 as the backbone network brings further improvement compared to using 3D R50, indicating that a better backbone network yields better detection performance.

\noindent\textbf{Visualization of the captured artifacts.} For a more in-depth understanding of how AltFreezing works, we further use Classification activation maps~(CAM)~\cite{cam} to localize which regions are activated to detect artifacts. The visualization results are shown in \cref{fig:cam}. Neither the vanilla 3D R50 nor the FTCN~\cite{FTCN} can notice the precise regions that are indeed manipulated. 3D R50 focuses on a very limited area for discrimination, which 
confirms that n\"iave training a 3D ConvNet leads to a trivial solution. FTCN~\cite{FTCN} pays more attention to locations outside of forged areas compared to 3D R50. In contrast, our AltFreezing makes it discriminates between real and fake by focusing predominantly on the manipulated face area. This visualization further identifies that AltFreezing encourages the 3D ConvNet to capture more spatial and temporal artifacts.

\vspace{1em}
\section{Conclusion and Discussion}
In this paper, we seek to capture both spatial and temporal artifacts in one model for more general face forgery detection. Concretely, we present a training strategy called AltFreezing that separates the spatial and temporal weights into two groups and alternately freezes one group of weights to encourage the model to capture both the spatial and temporal artifacts. Then, we propose a set of video-level fake data augmentations to encourage the model to capture a more
general representation of different  manipulation types.
Extensive experiments verify the effectiveness of the proposed AltFreezing training strategy and video-level data augmentations.
We hope that our work can attract more attention to video-level representation learning in the face forgery detection community.

\noindent\textbf{Acknowledgement.} This work was supported in part by the National Natural Science Foundation of China under Contract 61836011 and 62021001, and in part by the Fundamental Research Funds for the Central Universities under contract WK3490000007. It was also supported by the GPU cluster built by MCC Lab of Information Science and Technology Institution and the Supercomputing Center of USTC.

{\small
\bibliographystyle{ieee_fullname}
\bibliography{egbib}
}

\clearpage
\newpage
\appendix

\section{More Implementation Details}
\noindent\textbf{Face detection and align.} We use RetinaFace~\cite{Deng_2020_CVPR} to detect and align faces for each video. We crop the region of faces within the same range of the detected face area, \ie, four times the detected face area, where the weight and height are equal to twice the weight and height of the detected face, respectively. Then during training, for each video clip that contains 32 frames, we align them to a mean face. After all, the images in each clip are resized to 224 $\times$ 224.

\noindent\textbf{Network architecture.} The backbone we used is the bottleneck design of 3D ResNet50~(R50)~\cite{3dr50}, in which the $3\times 3$ convolution in the basic block is replaced with a consecutive $3\times 1 \times 1$ and $1\times 3 \times 3$ convolution. Our implementation is based on Pytorch 1.8.0 with Cuda 11.0 on 2 GeForce RTX 3090 GPUs.

\section{Additional Experiments}
\noindent\textbf{More ablation study of the freezing ratio  of AltFreezing.}
In our main paper, we have explained that adjusting the freezing ratio of $I_s:I_t$ can encourage the
network to pay more attention to spatial or temporal artifacts. And in default, we set the freezing ratio larger than 1. Here, we conduct more experiments to identify the effect of freezing ratio smaller than 1. The AUC results are reported in \cref{tab:alterratio_morespatial}. From the comparisons, we observe that  AltFreezing’s performance initially increases and then decreases as the freezing ratio varies from
1:1 to 1:20. The model achieves the best average AUC when the freezing ratio is 1:5, improving 0.4\% AUC compared to the baseline~(without AltFreezing) on average. Yet the performance is much lower than that when the freezing ratio $I_s:I_t$ is larger than 1. This is consistent with previous temporal-based methods~\cite{FTCN,haliassos2021lips} that claim detecting temporal artifacts is more general than detecting spatial ones.

\noindent\textbf{Ablation study of the fake clip generation.} To learn better video-level representation, we have proposed a set of fake video synthetic methods including temporal-level and spatial-level augmentations. We further conduct experiments to verify the effect of the components in the fake clip generation. The AUC results of the augmentations are reported in \cref{tab:video_aug_more_ablation}. We observe that enabled with the temporal augmentations~(ours~(w/o CB)), the model gets performance improvement on DFD~\cite{deepfakedetection} and FSh~\cite{li2020advancing}. On CDF it gets a performance drop. In our experiments, we use temporal augmentations in default since they might benefit the generalization ability to more challenging scenes. Moreover, clip-level blending which introduces more general clip-level spatial artifacts without any temporal artifacts further boosts the performance, averaging AUC from 95.6\% $\rightarrow$ 96.7\%.

\begin{table}[t]
\footnotesize
\centering
\resizebox{0.85\columnwidth}{!}{
\tablestyle{7pt}{1.08}
\begin{tabular}{ccccc}
\hline
Freezing & \multicolumn{3}{c}{Train on FF++} & \\ \cline{2-4} 
 ratio ($I_s:I_t$) &  FF++ & CDF & FSh & Avg\\ \hline  
 baseline & 99.3   & 81.8   & 99.2  &  93.4   \\ \hline
1:1 & 99.6 & \textbf{82.4} & 99.2 & 93.7  \\ 
1:5 & 99.7   & 82.2   & \textbf{99.4}   & \textbf{93.8} \\
1:20 & \textbf{99.8}   & 80.5   & 99.2   & 93.2 
 \\\hline
\end{tabular}
}
\vspace{-2mm}
\caption{\textbf{Ablation study of the ratio of freezing temporal kernels more than spatial ones of AltFreezing.} Video-level AUC(\%) is reported. ``baseline'' means a 3D R50 with end-to-end training.}
\vspace{-1.5em}
\label{tab:alterratio_morespatial}
\end{table}

\begin{table}[t]
\footnotesize
\centering
\begin{tabular}{cccccc}
\hline
\multirow{2}{*}{Aug.} & \multicolumn{4}{c}{Train on FF++} & \\ \cline{2-5} 
 &  FF++ & CDF & DFD & FSh & Avg\\ \hline   
 none & 99.7 & 86.4 & 97.6 & 99.3 & 95.8
 \\  
 ours~(w/o CB) & 99.7 & 84.5 & \textbf{98.8} & \textbf{99.4}& 95.6
 \\      
 \textbf{ours}& 99.7 &  \textbf{89.5}& 98.5 & 99.3 & \textbf{96.7}
 \\ \hline
\end{tabular}
\vspace{-2mm}
\caption{\textbf{Ablation study of the fake clip generation.} Video-level AUC(\%) is reported. ``Aug.'' means augmentation. ``CB'' denotes the clip-level blending in our fake clip generation.}
\label{tab:video_aug_more_ablation}
\vspace{-1em}
\end{table}

\section{Evaluation on Real-world Scenarios}
We further evaluate the performance of our model on more challenging scenes. The real-world DeepFake videos we used are downloaded from
the YouTube channel ``Ctrl Shift Face2''\footnote{\url{https://www.youtube.com/channel/UCKpH0CKltc73e4wh0_pgL3g}}, which are carefully crafted so that humans cannot discriminate between real and fake videos easily.
We compare our method with 3D R50~(baseline) without our AltFreezing and FTCN~\cite{FTCN}, as shown in the Youtube Url\footnote{\url{https://www.youtube.com/watch?v=q0m8r380P-A}}. Our method has a more
accurate judgment of real or fake. The comparison indicates that our method is much more robust than others in real-world scenarios.

\section{Limitations} We are aware that our method cannot handle any type of face forgery. When facing some fake videos generated by artists using Adobe Photoshop or other realistic image editing applications, our method may not be able to detect them. Besides, our method is not fully robust to all perturbations. For example, when applied to heavily compressed videos, the performance of our method drops  like other works.

\end{document}